\documentclass[letterpaper, 10 pt, conference]{ieeeconf}  

\IEEEoverridecommandlockouts                              

\usepackage{graphicx}
\usepackage{float}
\usepackage{amsmath}
\usepackage[section]{placeins}
\usepackage{lipsum} 
\usepackage{environ} 
\usepackage{amsmath}
\usepackage{tabularx}
\usepackage{adjustbox}
\usepackage[dvipsnames]{xcolor}
\usepackage{comment} 
\usepackage{setspace}
\usepackage{fixltx2e}
\usepackage{balance}

\title{\LARGE \bf
An Evaluation and Ranking of Different Voting Schemes for Improved Visual Place Recognition
}

\author{Maria Waheed$^{1}$,  Michael Milford$^{2}$, Xiaojun Zhai$^{1}$, Klaus McDonald-Maier$^{1}$ and Shoaib Ehsan$^{1,3}$
\thanks{*This research was supported by the UK Engineering and Physical Sciences Research Council (EPSRC) through grants EP/R02572X/1,  EP/P017487/1 and EP/V000462/1. This research was also partially supported by funding from ARC Laureate Fellowship FL210100156 to MM and the QUT Centre for Robotics. \textit{(Corresponding author: Maria Waheed)}}
\thanks{$^{1}$M. Waheed, X. Zhai, K. McDonald-Maier and S. Ehsan are with the School of Computer Science and Electronic Engineering, University of Essex, Colchester CO4 3SQ, United Kingdom.
        {\tt\small (e-mail: mw20987@essex.ac.uk; xzhai@essex.ac.uk; kdm@essex.ac.uk; sehsan@essex.ac.uk;  )}}%
\thanks{$^{3}$M. Milford is with the School of Electrical Engineering and Computer
Science, Queensland University of Technology, Brisbane, QLD 4000, Australia
        {\tt\small (e-mail: michael.milford@qut.edu.au)}}%
\thanks{$^{4}$Shoaib Ehsan is also with the School of Electronics and Computer Science, University of Southampton, Southampton, SO17 1BJ
        {\tt\small (e-mail: s.ehsan@soton.ac.uk)}}        
}

\begin{document}

\maketitle
\thispagestyle{empty}
\pagestyle{empty}

\begin{abstract}
Visual Place Recognition has recently seen a surge of endeavours utilizing different ensemble approaches to improve VPR performance. Ideas like multi-process fusion or switching involve combining different VPR techniques together, utilizing different strategies. One major aspect often common to many of these strategies is voting. Voting is widely used in many ensemble methods, so it is potentially a relevant subject to explore in terms of its application and significance for improving VPR performance. This paper attempts to looks into detail and analyze a variety of voting schemes to evaluate which voting technique is optimal for an ensemble VPR set up. We take inspiration from a variety of voting schemes that exist and are widely employed in other research fields such as politics and sociology. The idea is inspired by an observation that different voting methods result in different outcomes for the same type of data and each voting scheme is utilized for specific cases in different academic fields. Some of these voting schemes include Condorcet voting, Broda Count and Plurality voting. Voting employed in any aspect requires that a fair system be established, that outputs the best and most favourable results which in our case would involve improving VPR performance. We evaluate some of these voting techniques in a standardized testing of different VPR techniques, using a variety of VPR data sets. We aim to determine whether a single optimal voting scheme exists or, much like in other fields of research, the selection of a voting technique is relative to its application and environment. We also aim to propose a ranking of these different voting methods from best to worst according to our results as this will allow for better selection of voting schemes. 
\end{abstract}

\section{INTRODUCTION}
Efficiently performing the task of visual place recognition still remains an open field of research due to the extremely varying nature of the environments encountered [1]-[3]. A variety of VPR techniques exist each with its own merits and demerits but there is no universal technique equipped to perform equally well for all types of variations. This has led to the development of several ensemble methods [4]-[9], \begin{figure}[!htb]
    \centering
    \vspace*{0.1in}
    \includegraphics[width=1\columnwidth]{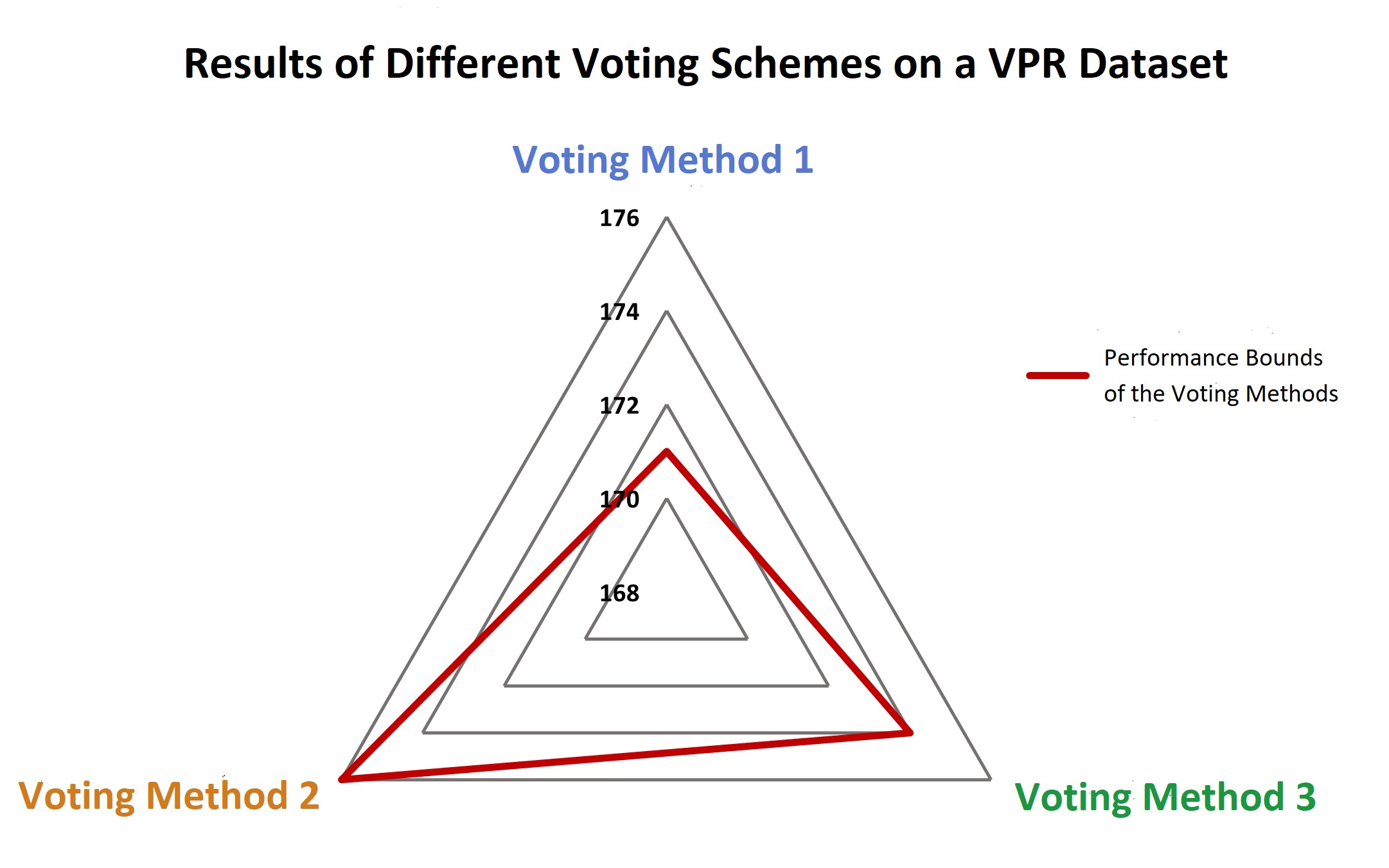}
    \caption{Sample output from the proposed experimental setup to evaluate the difference in performance bounds between each voting methodology. The same set of voting methods is tested for a single data set to anaylze the different results produced. The red line in the radar chart is representative of the performance bounds (images correctly matched) and the axis represent the total number of query images in the data set. The various dimensions are to useful to interpret the difference in performance of a voting scheme, in comparison to the others, such that the closer the red line is to boundary the better the performance of the voting scheme. For this given example the ranking in performance from best to worst is Voting method 2,3 and then 1.  }
    \label{fig:my_label}
\end{figure}one of which is multi-process fusion where [4],[5] work by selecting and combing different VPR  techniques to  match the query image from a sequence of images generated. Similarly, another idea proposed is complementarity among different VPR techniques presented by [6] while [7] switches and selects between different VPR techniques based on probability of correct match and complementarity. Other interesting work is also presented by [8],[9] where [8] attempts to gather a collection of very small CNN voting units to improve VPR performance. This is achieved by selecting the prediction of matched image that most units agree on. Finally, [9] discusses the idea of probabilistic voting for VPR utilizing nearest neighbor descriptor voting. A common phenomena among all these attempted ensemble approaches is that they employ one or another form of voting. Many researchers have attempted to create a uniform and standardized voting system that can be considered fair or optimal in any scenario. However, the exploration of different voting schemes to determine which is the optimal approach to use in an ensemble VPR set up is an area that has not yet been explored and can help produce some interesting insights as explained in Fig. 1.

\section{Methodology}
This section presents the selected voting schemes and their methodologies, that have been employed to a series of different VPR data sets to observe how the use of different voting mechanisms effects overall results in a basic ensemble VPR system. We test these voting schemes in a VPR set up that is simultaneously employing all state-of-the-art VPR techniques available and the final step involves selecting the correct reference image by using a voting methodology. 

The structure of our methodology section is based on analyzing different voting schemes that have been carefully selected to include unique voting systems to be tested. This analysis will provide us with the useful insights into what voting techniques are more useful for ensemble VPR methods and aid in ranking them from best to worst. This work will also help determine whether there is a clear winner when selecting the type of voting technique to employ or if its a relative choice dependant on other factors such as different data sets or type of ensemble method.

\subsection{Voting Scheme I : Plurality Voting}

This type of a voting mechanisms belongs to the family of positional voting which is a voting system that involves different ranks for different candidates and each rank holds a different priority. For plurality voting the candidate or option which has the most first place votes than any other candidate in the run is selected to be the final match. This is the most common or basic type of voting and is quite similar to hard voting that is used when trying to solve a classification problem. When dealing with an ensemble of VPR techniques each of which results in different reference images selected to match with the query image, it is not always simple or obvious which image is the correct match to the query. Here we present how plurality voting is employed to test out the results it would produce over different VPR data sets.

Let $i$ be the retrieved image and $v\textsubscript{i}$  is the number of votes each retrieved image acquires. $Argmax$ returns the reference image with the maximum votes, which is the final selected matched image to the query.

\begin{equation}
    \centering
    argmax \; i\in \{1,2,...,n\} \; v_{i}
\end{equation}

\subsection{Voting Scheme II : Condorcet Voting}
Condorcet is a ranked type of voting method that attempts to determine the overall selection of the reference image by comparing the results of each VPR technique from the ensemble techniques in one-on-one match-ups. Each VPR technique produces a sequence of potential correct images that are ranked on the basis of matching scores and a matrix is created with the number of times each image is ranked higher than the other possible correct matches. This matrix is used to determine the Condorcet winner. The reference image to win the most head-to-head match-ups is used to 
\begin{table}[]
\centering
\caption{VPR-Bench Datasets [26]} 
\begin{tabular}{|l|c|c|c|}
\hline
\textbf{Dataset}      & \multicolumn{1}{l|}{\textbf{Environment}} & \multicolumn{1}{l|}{\textbf{Query }} & \multicolumn{1}{l|}{\textbf{Ref. Images }} \\ \hline
GardensPoint & University                          & 200                                        & 200                                      \\ \hline
ESSEX3IN1   & University                          & 210                                        & 210                                      \\ \hline
CrossSeasons & City-Like                                 & 191                                        & 191                                      \\ \hline
Corridor     & Indoor                                    & 111                                        & 111                                      \\ \hline
17Places    & Indoor                                    & 406                                        & 406                                      \\ \hline
Livingroom   & Indoor                                    & 32                                         & 32                                       \\ \hline
\end{tabular}
\end{table} 
determine the winner based on the strength of their victories in the pairwise match-ups.

Let C be the set of references images where n is the total number of reference images. 
\begin{equation}
    \centering
     C = \{c_1, c_2, ..., c_n\}
\end{equation}

Let's denote the ranked positions as B where v is the total number of ranks. 
\begin{equation}
    \centering
     B = \{b_1, b_2, ..., b_v\}
\end{equation}

\begin{figure*}[!htb]
    \vspace*{0.1in}
    \includegraphics[width=2\columnwidth]{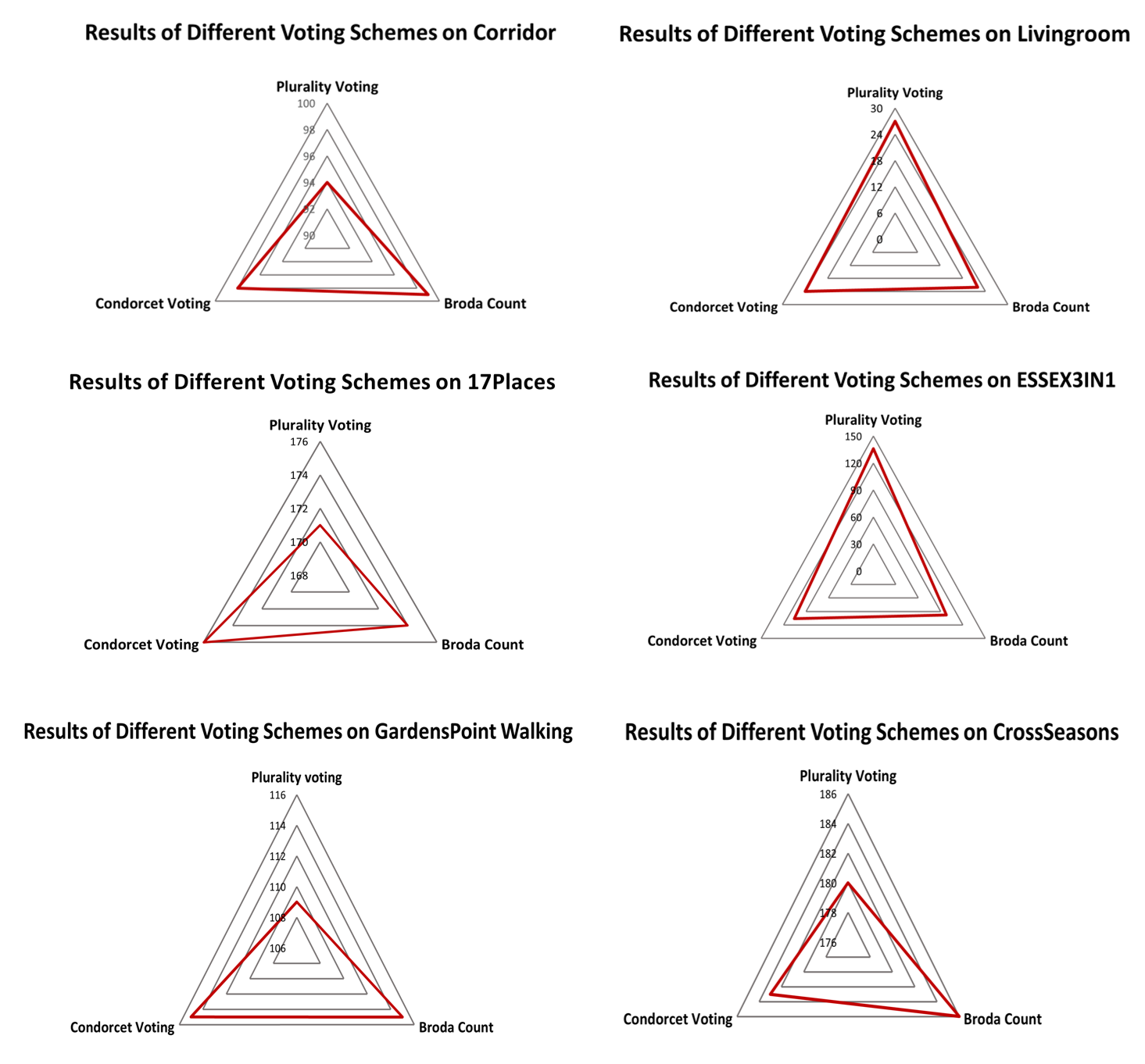}
   \caption{Difference in performance bounds of each voting methodology including Plurality, Condorcet and Broda Count voting, in terms of query images correctly matched for different data sets : Corridor (top left), Livingroom (top right), 17Places (center left), ESSEX3IN1 (center right), GardensPoint Walking (bottom left) and CrossSeasons (bottom right). 
   }
    \label{figurelabel}
\end{figure*}

Each \(b\textsubscript{v}\) in \(B\) contains a ranked preference order for the reference images in C. For example, if we have 3 references images \(c\textsubscript{1}\),\(c\textsubscript{2}\),\(c\textsubscript{3}\) a ranked positions could be \(b\textsubscript{v} (c\textsubscript{1},c\textsubscript{2},c\textsubscript{3})\), indicating that \(c\textsubscript{1}\) is the technique’s first choice, \(c\textsubscript{2}\) is the second choice, and \(c\textsubscript{3}\) is the third choice. The Condorcet winner is the reference image that would win in a head-to-head matchup against any other candidate. To calculate the pairwise victories for each pair of candidates \(c\textsubscript{i}\) and \(c\textsubscript{j}\) in C, we add the number of times \(c\textsubscript{i}\) is ranked higher than \(c\textsubscript{j}\). The reference image that is ranked higher than all other images is the final selected reference image.

\subsection{Voting Scheme III : Broda Count Voting}
Broda Count is another type of ranked or positional voting system utilized for various types of electoral tasks. Much like all other voting methodologies it  has its own strengths and weakness but it can be useful to explore the type of results it can produce for different ensemble VPR techniques. Similar to the plurality voting it also belongs to the family of positional voting where the candidates or potential matches to the query image are ranked in a descending order based on their matching scores. The position or rank of the reference image is important as a higher rank (i.e higher points/score) suggests a higher chance or preference of the particular image being selected as the final match in the ensemble VPR set up.

Let \(a\) be the reference image selected while \(i\) represents the total number of techniques being considered. \(j\) represents the rank of each reference image and $n$ is the value of points or score for a given rank. And finally \(S\textsubscript{a}\) is the summation of all the points for a single reference image. 
\begin{equation}
    \centering
    S_{a} = \sum_{i=1}^{i=n} x_{i,j}
\end{equation}
The image with the highest sum is selected as the final match to the query. 
\begin{equation}
    \centering
    argmax \; i\in \{1,2,...,n\} \; S_{a}
\end{equation}

\section{Experimental Setup}

The evaluation of these different voting methodologies help observe how a voting scheme can have significantly different results than another for the same type of ensemble VPR set up. For this experiment different VPR data sets have been utilzied to test this idea and are presented in TABLE I lists all the data sets including Corridor [10], Living room [11], ESSEX3IN1 [12], GardensPoint [13], Cross-Seasons [14], and 17Places [15], along with their environmental set up, number of query and reference images. Eight state-of-the-art VPR techniques have been selected for the experimentation, to test the different voting schemes on an ensemble set up and these include AMOSNet [16], HOG [17],[18], and  AlexNet [19],[20], HybridNet [21], NetVLAD [22], RegionVLAD [23], CoHOG [24], and CALC [25].

\section{Results} 
In this section we present the results produced by testing each of the selected voting methodologies over various VPR datasets, under the same ensemble VPR set up. The results presented are in terms of the sum of query images correctly matched employing each of the different voting schemes. Results of six widely employed VPR data sets are discussed and presented in the form of a radar chart displaying the difference in performance bounds of each voting scheme. The ideal scenario would be to identify a single voting method that performs better than all the others present. Fig. 2 presents the results that were produced starting from Corridor and CrossSeasons data set, and their ranking from best to worst is Broda Count, Condorcet voting and lastly, plurality voting. For data sets Livingroom and ESSEX3IN1 the results are quite different as their ranking appears to be Plurality, Condorcet and then Broda Count voting. Finally, Gardenspoint and 17Places data sets have plurality voting as their the least favourable option while, Broda count voting is second and the best option according to the results appears to be Condorcet voting for the two data sets. It is interesting to note that even though there is no one voting scheme that seems to be optimal for each data set, there appears to be a uniform distribution among the results of these six data sets, with two data sets favoured by each voting scheme. This helps us conclude that each of these voting methods holds some merit and has further potential for exploration on other VPR data sets under a different environmental set up for a more conclusive result. 

\section{Conclusion}
In this work we propose to explore the different kinds of voting schemes that can be used in an ensemble VPR set up and the performance difference between them, under same experimental set up. Although a variety of voting schemes exists in different fields of research, for this work we have selected a total of three different voting methods to test including Plurality voting, Condorcet voting and Broda count voting. Although the results did not determine a single voting method as the the best voting scheme among the three, they do however help conclude that each of these voting scheme is viable in different cases and further exploration should be attempted. Additionally, we have evaluated results in terms of correctly matched images per data set but it quite likely that different and more nuanced results can be produced when considering other evaluation methods such as recall@X. It might be an interesting path forward to explore other voting methods such as instant run-off voting, Contingent Voting and Approval voting etc on a more extensive experimental set up with a different evaluation matrix to determine and further rank the voting schemes from best to worst.

\end{document}